\documentclass[letterpaper, 10 pt, conference]{ieeeconf}  

\IEEEoverridecommandlockouts                              

\usepackage{graphics} 
\usepackage{graphicx}

\usepackage{epsfig} 
\usepackage{amsmath} 
\usepackage{amssymb}  

\usepackage{bm}
\usepackage{mathtools}

\usepackage{algorithm}
\usepackage[noend]{algpseudocode} 
\algnewcommand\AAND{\textbf{ and }}
\algnewcommand\Or{\textbf{ or }}

\usepackage{color}
\usepackage{citesort}
\usepackage{flushend}
\usepackage{url}
\usepackage[breaklinks]{hyperref}
\usepackage[capitalise]{cleveref}
\usepackage{booktabs}
\usepackage{makecell}
\usepackage{multirow}

\usepackage[nolist,nohyperlinks]{acronym}
\acrodef{method}[AOM]{ACRONYM OF METHOD}
\acrodef{gnss}[GNSS]{Global Navigation Satellite System}
\acrodef{ransac}[RANSAC]{Random Sample Consensus}
\acrodef{slam}[SLAM]{Simultaneous Localization And Mapping}
\acrodef{pca}[PCA]{Principal Component Analysis}
\acrodef{ekf}[EKF]{Extended Kalman Filter}
\acrodef{rmse}[RMSE]{Root Mean Square Error} 
\acrodef{ape}[APE]{Absolute Pose Error}
\acrodef{cfar}[CFAR]{Constant False Alarm Rate}
\acrodef{snr}[SNR]{Signal to Noise Ratio}
\acrodef{rcs}[RCS]{Radar Cross Section}
\acrodef{imu}[IMU]{Inertial Measurement Unit}
\acrodef{sgm}[SGM]{Segmi-Global Matching}
\acrodef{dnn}[DNN]{Deep Neural Network}
\acrodef{gru}[GRU]{Gated Recurrent Unit}
\acrodef{hpr}[HPR]{Hidden Point Removal}
\acrodef{raft}[RAFT]{Recurrent All-Pairs Field Transforms}
\acrodef{fov}[FOV]{Field of View}
\acrodef{mclab}[MC-lab]{Marine Cybernetics laboratory}
\acrodef{vio}[VIO]{Visual-Inertial Odometry}
\acrodef{vae}[VAE]{Variational Autoencoder}
\acrodef{dce}[DCE]{Deep Collision Encoder}
\acrodef{drl}[DRL]{Deep Reinforcement Learning}
\acrodef{lmf}[LMF]{Learning-based Micro Flyer}

\usepackage{tikz}
\usetikzlibrary{positioning, shapes.geometric, arrows.meta, decorations.pathreplacing, calligraphy}

\DeclareMathAlphabet{\pazocal}{OMS}{zplm}{m}{n}

\newcommand{\Bs}{\pazocal{B}}

\newcommand{\Ls}{\pazocal{L}}
\newcommand{\Vs}{\pazocal{V}}

\newcommand{\Ss}{\pazocal{S}}

\newcommand{\Is}{\pazocal{I}}

\DeclareMathAlphabet{\mathpzc}{OT1}{pzc}{m}{it}

\usepackage{array}
\newcolumntype{C}[1]{>{\centering\arraybackslash}p{#1}}
\newcolumntype{M}[1]{>{\raggedright\arraybackslash}p{#1}}

\usepackage{array} 
\newcolumntype{L}[1]{>{\raggedright\let\newline\\\arraybackslash\hspace{0pt}}m{#1}}	
\newcolumntype{S}[1]{>{\centering\let\newline\\\arraybackslash\hspace{0pt}}m{#1}}
\newcolumntype{R}[1]{>{\raggedleft\let\newline\\\arraybackslash\hspace{0pt}}m{#1}}

\newcommand{\x}{\mathbf{x}}
\newcommand{\xcoll}{\mathbf{x}_{\textrm{coll}}}

\makeatletter
\renewcommand*{\@opargbegintheorem}[3]{\trivlist
  \item[\hskip \labelsep{\itshape #1\ #2}] \textit{(#3)}\ }
\makeatother

\title{\LARGE \bf
Reinforcement Learning for Collision-free Flight Exploiting \\ Deep Collision Encoding
}

\author{Mihir Kulkarni and Kostas Alexis 
\thanks{This work was supported by the AFOSR Award No. FA8655-21-1-7033 and the Horizon Europe Grant Agreement No. 101070405.}
\thanks{The authors are with the Autonomous Robots Lab, Norwegian University of Science and Technology (NTNU), O. S. Bragstads Plass 2D, 7034, Trondheim, Norway {\tt\small mihir.kulkarni@ntnu.no}}
}

\begin{document}

\maketitle
\thispagestyle{empty}
\pagestyle{empty}

\begin{abstract}

This work contributes a novel deep navigation policy that enables collision-free flight of aerial robots based on a modular approach exploiting deep collision encoding and reinforcement learning. The proposed solution builds upon a deep collision encoder that is trained on both simulated and real depth images using supervised learning such that it compresses the high-dimensional depth data to a low-dimensional latent space encoding collision information while accounting for the robot size. This compressed encoding is combined with an estimate of the robot's odometry and the desired target location to train a deep reinforcement learning navigation policy that offers low-latency computation and robust sim2real performance. A set of simulation and experimental studies in diverse environments are conducted and demonstrate the efficiency of the emerged behavior and its resilience in real-life deployments. 

\end{abstract}

\section{Introduction}\label{sec:intro}
Aerial robots are tasked to undertake ever more complex missions in demanding environments, including high-risk applications in GPS-denied, cluttered environments such as subterranean exploration~\cite{CerberusScience,CERBERUS_WINS_FR2022submission,CERBERUS_SUBT_PHASE_I_II}, forest under canopy mapping~\cite{zhou2022swarm}, industrial inspection in confined facilities~\cite{thakur2020nuclear,ozaslan2017autonomous}, and search and rescue missions~\cite{delmerico2019current}. Key to enabling resilient autonomy is identifying core functionalities that experience significant impediments in their performance and designing novel approaches to overcome such limitations. A prevalent issue especially for small flying robots is the typically-employed separation of collision-free motion planning and control, with the first relying on (online) maps enabling collision checking and the second merely following the commanded paths. As maps can present errors or not capture certain obstacles, this approach is prone to failures as discussed in the framework of challenging deployments like the DARPA Subterranean Challenge~\cite{CerberusScience,ebadi2022present}. Furthermore, collision-free flight that requires the online reconstruction of maps is bound to involve high-latency operations reducing the maximum possible update rate. 

\begin{figure}[ht]
\centering
    \includegraphics[width=0.972\columnwidth]{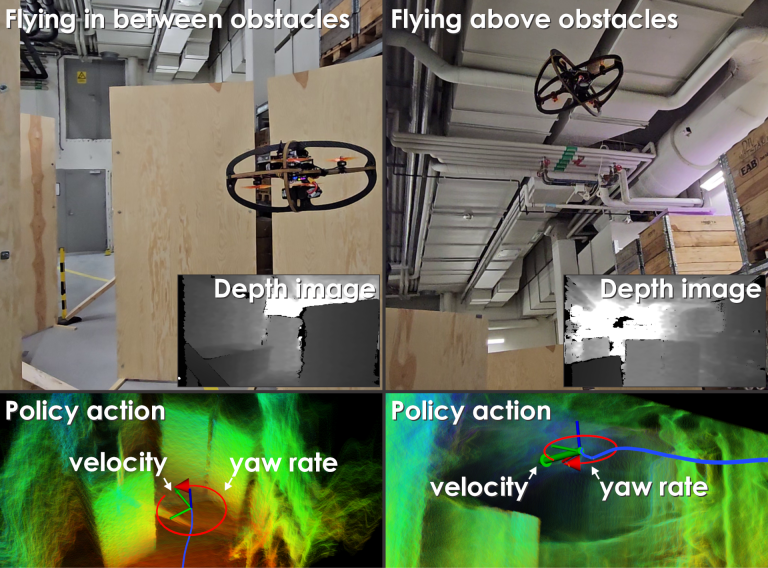}
\vspace{-2.5ex}
\caption{Instances of two experiments demonstrating the abilities of the navigation policy trained using deep collision encoding and trained with reinforcement learning. If allowed and space is available, the intuitive behavior of flying above all obstacles is selected (right), while when the robot is constrained regarding its altitude it is capable of maneuvering through highly cluttered settings (left).}
\label{fig:vaerl_intro}
\vspace{-4ex}
\end{figure}

Motivated by the above, in this work we develop a novel navigation policy that learns to enable collision-free flight in confined environments, while solely relying on a single real-time depth observation and an estimate of the robot's odometry. Focusing on robust sim2real transfer, a modularized deep learning solution is proposed. Specifically, the method first exploits a \ac{dnn} that serves as a \ac{dce} using the high-dimensional real-time depth image data as input and compressing them to a very low-dimensional latent space that retains information for collision building upon the principles of \acp{vae}. In particular, depth images with resolution equal to $640\times 480$ pixels are compressed to a latent vector with only $64$ dimensions ($4800\times$ compression). Given the inference of this latent space in real-time, a \ac{drl} navigation policy is trained that exploits the latent space and the estimate of the robot's state, alongside information for the target location to enable an aerial robot to fly safely within cluttered environments by commanding reference velocities and yaw rate to the system's low-level autopilot. 

The proposed deep learning navigation policy boasts a set of contributions. First, through its modular architecture and the \ac{dce} it allows to reduce the sim2real gap as on one hand the validity of its low-dimensional latent space, trained with supervised learning, can be verified separately and on the other it allows to assimilate real-life depth image data for training. Second, it represents a \ac{drl} policy for aerial robots to fly through diverse cluttered environments without being limited to a particular task and associated environment assumptions or knowledge, such as for example in the recent pioneering work in drone racing~\cite{kaufmann2023champion,song2021autonomous,yunlog2023reaching}. Third, it represents a method verified not only in simulation but also on challenging experimental studies such as those in Figure~\ref{fig:vaerl_intro}, while further ensuring that low-latency inference is achieved onboard. We discuss the emerged behaviors, such as increasing or decreasing speed as necessary to maneuver around obstacles or flying above them all when possible. 

In the remaining paper, Section~\ref{sec:related} presents related work, followed by the problem statement in~\ref{sec:problem}. The proposed approach is detailed in Section~\ref{sec:approach}, with evaluation studies in Section~\ref{sec:evaluation} and conclusions in Section~\ref{sec:concl}.


\section{Related Work}\label{sec:related}
Deep learning has recently evolved as a focal point of research in collision-free navigation. A subset of this research aims to solve the global planning problem, where a complete map of the environment is available beforehand. These approaches may use top-down images or point clouds of the whole environment to plan paths~\cite{Brian2019Latent,Srinivas2018UPN,Qureshi2021MPN}. Contrary to this line of work, here we focus on local navigation exploiting solely onboard observations without access to a global map. 

In the domain of local navigation, a set of studies use imitation learning techniques to generate collision-free trajectories, which are then followed by model-based controllers~\cite{Loquercio2021Science,Tolani2021visual}. Methods that learn over different action spaces (velocity/steering angle, acceleration, or angular velocity/thrust commands) have also been explored, for example using a) reinforcement learning~\cite{Francis2020PRMRL,ugurlu2022sim}, b) supervised learning where ground-truth commands are readily available in a driving dataset~\cite{Loquercio2018Dronet}, provided by human operators~\cite{shah2022viking} or demonstrated by an expert~\cite{kaufmann2020RSS}, as well as c) self-supervised learning~\cite{Gandhi2017learning2fly,Kahn2021BADGR,Kahn2021LAND}. In~\cite{ugurlu2022sim}, the authors employ an end-to-end approach for training a \ac{drl} navigation policy that exploits depth data but train without considering the dynamics of micro aerial vehicles and test in sparse environments. 

Further research in local autonomous navigation employs deep learning to create interpretable maps. These maps are then used by classical planners to navigate without collisions~\cite{Wang2021Occluded,Frey2022Locomotion,Castro2022How,Zeng2019Interpretable}. Alternatively, some works bypass traditional map representations and directly encode raw sensor data into latent vectors. These vectors are then used to infer control actions, enabling low-latency navigation~\cite{Hoeller2021representation,Dugas2021Navrep,Brian2019Latent,Srinivas2018UPN,Qureshi2021MPN,Loquercio2021Science,grando2020deep,grando2022double}. Previous work of the authors has explored the use of supervised deep collision prediction allowing to classify which among a set of motion primitives collide or not with the environment~\cite{ORACLE,sevae_oracle}. Focusing on a particular application niche, namely drone racing, the authors in~\cite{kaufmann2023champion} build upon a host of investigations~\cite{Wagter2021AlphaDrone,Foehn2021AlphaDrone,song2021autonomous,loquercio2019deep,kaufmann2019beauty} and demonstrate beyond-human performance using deep reinforcement learning that exploits the particular structure of the problem (e.g., flying through known types of gates). 


\section{Problem Formulation}\label{sec:problem}
The problem considered in this work is that of autonomous collision-free aerial robot navigation assuming no access to the maps of the environment, neither from offline data nor online reconstruction, and with access only to a) an estimate of the robot's pose, alongside b) the immediate depth observation using a frustum- and range-constrained sensor. %
Let $\Is$, $\Bs, \Vs$ be the inertial, body- and vehicle ($\Is$ rotated to have the same yaw as the robot) frames respectively, $\mathbf{x}_t$ the current depth image from an onboard sensor, and $\mathbf{s}_t = [\mathbf{p}_t, \mathbf{v}_t, \mathbf{q}_t, \bm{\omega}_t]$ the estimated robot state consisting of a) its 3D location in $\Is$ ($\mathbf{p}_t = [{p}_{t,x},{p}_{t,y},{p}_{t,z}]$), b) the 3D velocity in $\Bs$ ($\mathbf{v}_t=[{v}_{t,x}, {v}_{t,y}, {v}_{t,z}] \in \mathbb{R}^{3 \times 1}$), c) the attitude here represented in quaternion form $\mathbf{q}$, and d) the angular velocity in $\Bs$ ($\bm{\omega}_t$). Given a 3D goal location $\mathbf{p}^{r}$ expressed in $\Is$, the problem is that of finding an optimized action vector $\mathbf{u}_t=[v_{t,x}^r,v_{t,y}^r,v_{t,z}^r,\omega_{t,z}^r]^T$ (in compact form $\mathbf{u}_t = [{\mathbf{v}_t^r}, \omega_{t,z}^r]^T$) involving the commanded robot velocities expressed on $\Bs$ and the yaw rate ${\omega}_{t,z}^r$ that can enable safe flight avoiding collisions despite the presence of a set of obstacles $\Ss_O$. This action vector $\mathbf{u}_t$ is then provided to be tracked by a low-level controller of the flying vehicle.


\section{Proposed Approach}\label{sec:approach}
The proposed approach on a deep learned modular collision-free navigation policy exploiting deep collision encoding and reinforcement learning is presented.

\subsection{Task-driven Compression for Collision Encoding}

At the core of the architecture of the proposed solution is the modularization of the two demanding learning tasks, namely a) processing high-dimensional depth image data to enable collision avoidance, and b) deriving an optimized policy for collision free navigation. In the first step, task-driven depth image compression for collision encoding takes place. In particular, building upon our work in~\cite{dce_isvc_2023}, a \ac{dnn} and specifically the architecture of a \ac{vae} is employed with the goal of encoding a high-resolution depth image to a very low dimensional latent space trained to retain the information necessary for collision prediction. This is a key distinction compared to methods that will attempt to train a method end-to-end which does not allow to assimilate real image data in the training in a practical way~\cite{ORACLE} and only allows to verify a navigation method regarding its ultimate performance with limited ability to ensure that separately verify how the high-dimensional image data are processed in a manner that ensures that collision information is retained.

\begin{figure*}[ht!]
\centering
\includegraphics[width=0.90\textwidth]{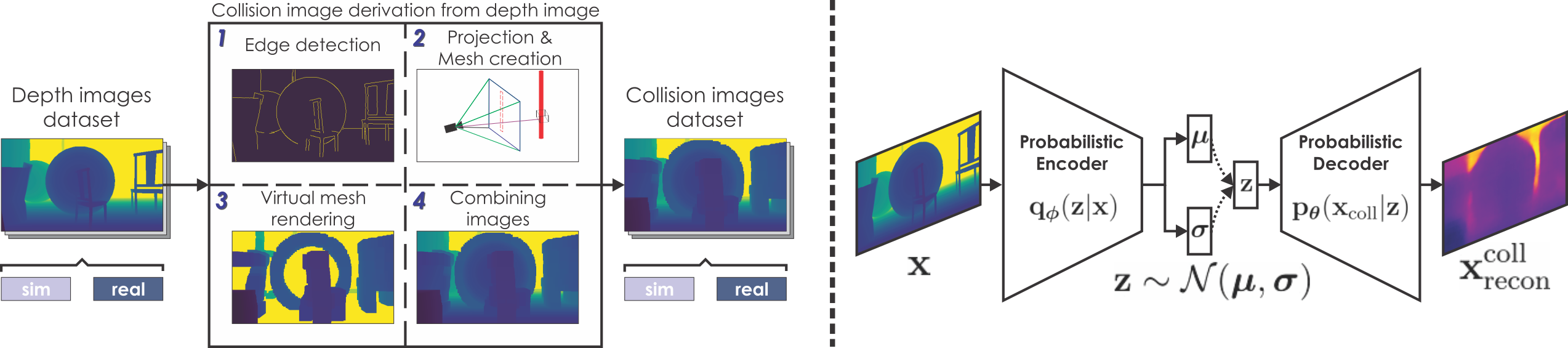}
\vspace{-2ex}
\caption{Overview of the Deep Collision Encoder used to derive a low-dimensional latent space that retains collision information from depth images. The \ac{dce} is trained using supervised learning that exploits a dataset involving both synthetic and real depth images. The depth images are transformed to collision images that account for the size of the robot. The involved \ac{dnn} exploits an architecture motivated by variational autoencoders, while the ``encoder'' and ``decoder'' elements are in fact also functioning to encode the depth image to the collision image and its reconstruction from the latent space. }
\label{fig:dcevae}
\vspace{-3ex}
\end{figure*}

In further detail, the proposed \ac{dce} considers depth images $\mathbf{x}$ that are remapped to derive a ``collision image'' $\xcoll$ that accounts for the fact that any $3\textrm{D}$ location captured within the sensor's frustum represents a collision-free point only if the robot with its non-zero size (modeled as a box with dimensions $D_R \times W_R \times H_R$) can fit in the respective $3\textrm{D}$ location. Through the \ac{dce}'s remapping-encoding step, followed by its decoding step, depth images are compressed to a low-dimensional latent space $\mathbf{z}$ that gives rise to the reconstructed $\mathbf{x}_{\textrm{recon}}^{\textrm{coll}}$ that closely resembles $\xcoll$ as detailed in~\cite{dce_isvc_2023} and summarized in Figure~\ref{fig:dcevae}. Of particular importance for the \ac{drl} navigation policy, the latent dimensions can be of very low dimension as $\xcoll$ is typically less complex than $\mathbf{x}$ owing to the ``inflation'' by the robot size. 

To learn to both compress and remap the depth image $\mathbf{x}$ such that the difference between the reconstructed $\mathbf{x}_{\textrm{recon}}^{\textrm{coll}}$ and the collision image $\xcoll$ is minimized, the loss function guiding the supervised training of the \ac{dce} takes the form:

\begin{align}
    \Ls &= \Ls_{\textrm{recon}} + \beta_{\textrm{norm}}\Ls_{KL},
\end{align}
where
\begin{align*}
    \Ls_{\textrm{recon}}(\xcoll, \x^{\textrm{coll}}_{\textrm{recon}}) &= \textrm{MSE}(\xcoll, \x^{\textrm{coll}}_{\textrm{recon}}),\\
    \Ls_{KL}(\bm{\mu}, \bm{\sigma}) &= -\frac{1}{2}\sum_{j=1}^J \left (1 + \log(\bm{\sigma}_j^2) - \bm{\mu}_j^2 - \bm{\sigma}_j^2 \right).
\end{align*}
Here, $\Ls$ denotes the overall loss consisting of the reconstruction loss $\Ls_{\textrm{recon}}$ and KL-divergence loss $\Ls_{KL}$, scaled by a constant $\beta_{\textrm{norm}}$~\cite{higgins2017betavae}. These terms are inspired by autoencoder literature~\cite{doersch2016tutorial}. MSE denotes Mean-Square Error loss terms. 

It is noted that focusing on sim2real transfer, in this work the method is extended in its ability to assimilate real data from a depth sensor despite the presence of invalid pixels driven by common stereo disparity challenges such as gaps, noise and quantization~\cite{nguyen2012modeling,fankhauser2015kinect,korkalo2021measurement}. To enable this goal, the contribution of invalid pixels is removed from the loss terms presented above. Furthermore, Bernoulli sampling is used to determine invalid pixels and noisy depth values are obtained by sampling from a normal distribution.


\subsection{Navigation Policy Learning}
We employ a \ac{drl} framework to train  neural network policies to navigate cluttered environments. We formulate this problem as a partially observable Markov decision process (POMDP) with the true state of the environment, robot and the previous action denoted by $\mathfrak{s}_t$ at a discrete time $t$. An action $\mathbf{a}_t \in \mathbf{A}$ can be applied to the environment to obtain a new state at the next time step $t+1$ with the transition probability $\mathbf{Pr}(\mathfrak{s}_{t+1} | \mathfrak{s}_{t}, \mathbf{a}_t)$. At each step, the agent observes the environment with the probability $\mathbf{O}(\mathbf{o}_{t+1} | \mathfrak{s}_{t+1})$. We utilize the robot state $\mathbf{s}_t$, goal position $\mathbf{p}^{r}$, and the depth image $\mathbf{x}_t$ to compute the observations that the agent gathers from the environment. The observations for the agent are defined as $\mathbf{o}_t = [ \hat{\mathbf{n}^{g}_{t}}, {||\mathbf{n}^{g}_{t}||}_{2},\mathbf{v}_t, \phi_t, \theta_t, \bm{\omega}_t, \mathbf{a}_{t-1}, \mathbf{z}_t]$, where $\hat{\mathbf{n}^{g}_{t}}$ denotes the unit vector from the robot position $\mathbf{p}_t$ to the goal position $\mathbf{p}^{r}$ expressed in the vehicle frame $\Vs$, ${||\mathbf{n}^{g}_{t}||}_{2}$ is the Euclidean distance between the robot and the goal, $\mathbf{v}_t$ and $\bm{\omega}_t$ are the linear and angular velocities respectively expressed in the body frame $\Bs$. $\phi_t$ and $\theta_t$ correspond to the current roll and pitch angles of the robot and $\mathbf{a}_{t-1}$ is the vector of actions obtained from the network at time $t-1$. Finally, $\mathbf{z}_t$ denotes the compressed latent representation of image $\mathbf{x}_t$ obtained from the \ac{dce}.

The agent also maintains a belief over the states where $\mathbf{B}_{t}(\mathfrak{s}_t)$ denotes the probability that the environment/robot is in state $\mathfrak{s}_t$. The belief is updated based on the current observation, the current action and the previous belief state as $\mathbf{B}_{t+1} = \bm{\tau}(\mathbf{B}_t, \mathbf{a}_t, \mathbf{o}_{t+1})$. For each state transition a reward is provided to the agent in the form of $\mathbf{R}(\mathfrak{s}_{t+1} | \mathfrak{s}_t, \mathbf{a}_t)$. A policy that yields actions given observations $\mathbf{a}_t = \bm{\pi}(\mathbf{o}_t)$ is learned by the agent such that the sum of rewards over an episode is maximized. The reward function for the state transitions is defined as:


\begin{align}
    \mathbf{R}(\mathfrak{s}_{t+1}|\mathfrak{s}_{t}, \mathbf{a}_{\textrm{t}}) &= \sum_{i=1}^{4} \lambda_i r_i + \sum_{j=1}^{2} \eta_j p_j + p_{\textrm{crash}},
\end{align}
where $r_i$ indicates a positive reward term and $p_i$ indicates a negative reward term, defined as:

\small
\begin{align*}
    r_1 &= r({||\mathbf{n}^{g}_{t}||}_{2}, \nu_1), \\
    r_2 &= r({||\mathbf{n}^{g}_{t}||}_{2}, \nu_2), \\
    r_3 &= \frac{|\nu_3 - {||\mathbf{n}^{g}_{t}||}_{2}|}{\nu_3}, \\
    r_4 &= \nu_4({||\mathbf{n}^{g}_{t}||}_{2} - {||\mathbf{n}^{g}_{t-1}||}_{2}), \\
    p_1 &= \sum^{}_{k}\nu_{5, k}(r(g(\mathbf{a}_{t})_k, \nu_{6, k})-1),\\
    p_2 &= \sum^{}_{k}\nu_{7, k}(r(g(\mathbf{a}_{t})_k - g(\mathbf{a}_{t-1})_k, \nu_{8, k})-1),\\
    p_{\textrm{crash}} &= -\nu_{9}.
\end{align*}
\normalsize
The function $r$ is defined as $r(x, \nu) = e^{-\frac{x^2}{\nu}}$. The values $\lambda_i, \eta_{j}>0$ and $\nu_{i}, \nu_{m,k}>0$ represent tuning parameters. $\nu_{1} \hdots \nu_{4}$ and $\nu_{9}$ are scalars while $\bm{\nu}_{5} \hdots \bm{\nu}_{8}$ are vectors with the same number of dimensions as the output of function $g$ as defined in the next subsection. These values may vary during training as discussed later in this Section. With this environment definition, we utilize the Asynchronous Proximal Policy Optimization (APPO) algorithm from Sample Factory~\cite{petrenko2020sample} to train a deep neural network policy to navigate a robot to the goal location in a collision-free manner.



\subsection{Implementation Details}
We define a neural network architecture containing $3$ fully-connected layers consisting of $512$, $256$ and $64$ neurons each with an ELU activation layer, followed by a GRU with a hidden layer size of $64$. Given an observation vector $\mathbf{o}_t$, the policy outputs a $3$-dimensional action command $\mathbf{a}_t = [a_{t,1}, a_{t,2}, a_{t,3}]$ with values in [-1, 1]. These action values are mapped to the speed, inclination of the commanded velocity with the $x$-axis of the robot and yaw-rate. A function $g$ is defined such that $\mathbf{u}_t = g(\mathbf{a_t})$, to convert these values to the input command for the velocity controller $\mathbf{u}_t=[v_{t,x}^r,v_{t,y}^r,v_{t,z}^r,\omega_{t,z}^r]^T$. The action outputs are converted to velocity and yaw-rate command as:
\begin{align*}
    v_{t,x}^r &= s_{\textrm{max}}  (\frac{a_{t,1} + 1}{2} \textrm{cos}(i_{\textrm{max}}a_{t,2})),\\
    v_{t,y}^r &= 0.0, \\
    v_{t,z}^r &= s_{\textrm{max}} (\frac{a_{t,1} + 1}{2}\textrm{sin}(i_{\textrm{max}}a_{t,2})),\\
    \omega_{t,z}^r &= \omega_{\textrm{max}} a_{t,3},
\end{align*}
where $s_{\textrm{max}}$ is the maximum speed and $i_{\textrm{max}}$ is the maximum angle of the commanded velocity with the $x$-axis of the robot, while $\omega_{\textrm{max}}$ is the maximum commanded yaw-rate. This parameterization of the controller commands and $s_{max},~i_{max},~\omega_{max}$ are chosen to ensure that commanded velocity vector lies within the field-of-view of the depth sensors and prevent a sideways collision with the environment. The neural network is trained with an adaptive learning rate initialized at $l_r = {10}^{-4}$. The discount factor is set to $\gamma=0.98$. The neural network is trained with $1024$ environments simulated in parallel with an average time step of $0.1\textrm{s}$ and rollout buffer size set to $32$. We train this policy for approximately $26\times10^6$ environment steps aggregated over all agents.

The \ac{dce} is pre-trained and frozen during the training of the RL policy. For training the \ac{dce} network, collision images are generated for $10,000$ real images and $21,000$ simulated images. The network architecture and training methodology followed is similar to~\cite{dce_isvc_2023}, with $\beta_{\textrm{norm}} = 3.0$. Furthermore, in this work, we utilize the sampled latent representation $\mathbf{z} \sim \mathcal{N}(\bm{\mu}|\bm{\sigma})$ for training the RL policy.

\subsection{Training Environment}

\begin{figure}[ht!]
\centering
\includegraphics[width=0.99\columnwidth]{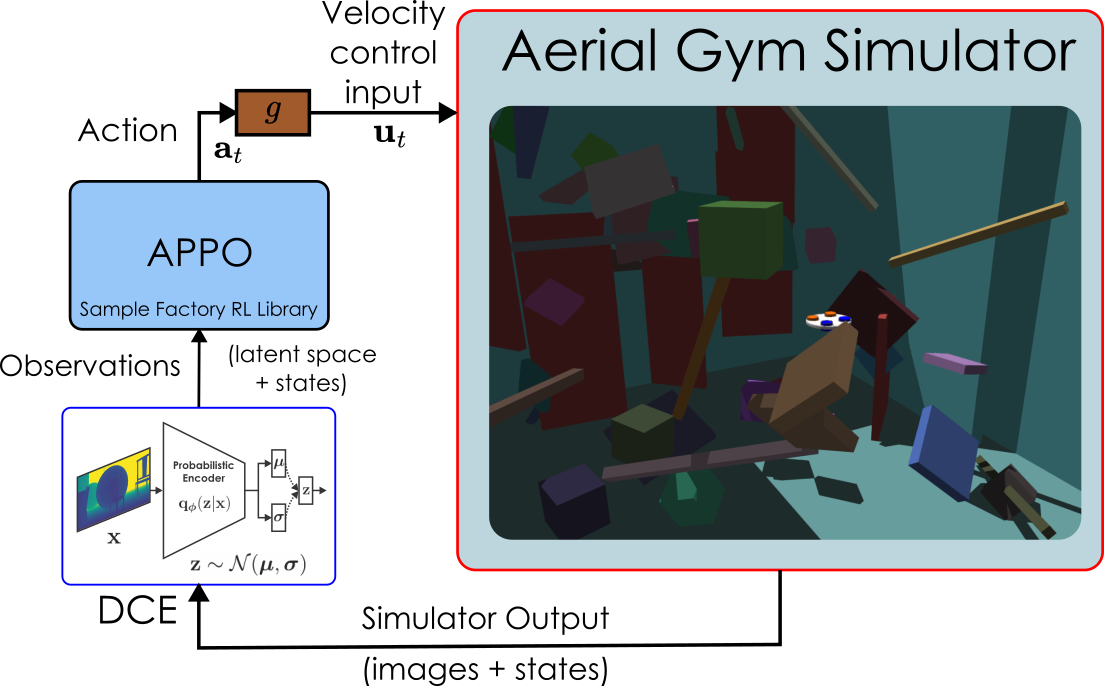}
\vspace{-2ex}
\caption{Overview of the interface between the framework for RL agent training, the Aerial Gym Simulator and the \ac{dce}.}
\label{fig:rl_training_diagram}
\vspace{-4ex}
\end{figure}

The Aerial Gym Simulator~\cite{kulkarni2023aerial} provides the environment and the interfaces to train the \ac{drl} policy to navigate cluttered environments. The \ac{dce} and the learning framework are interfaced with the simulator as shown in Figure~\ref{fig:rl_training_diagram}. The simulator provides capabilities for massively parallelized simulation of aerial robots that are equipped with depth cameras. The provided velocity controller is utilized for this work. We generate cluttered environments within the simulator consisting of a room-like environment bounded by walls containing static obstacles that are positioned and oriented in a randomized manner by uniformly sampling the position and Euler angles for each obstacles. These obstacles are kept floating in the environment (i.e., the effect of gravity is disabled on these objects) to allow more randomization in the training environments. While the length of an episode is fixed, each environment can run episodes asynchronously allowing different environments to provide the learning agent experience from different stages of the task simultaneously.

A curriculum learning framework is set up that gradually complexifies the environment by adding more objects to it as the agent learns to successfully navigate the robot in easier environments (with success rate greater than $70\%$), while making the environment easier if robots are crashing at a rate above a given threshold (here $30\%$). We set up a curriculum that involves logging the runs of the robots to measure successes, crashes and episode timeouts. Successful runs are said to happen when the robot is within a specified distance of the goal location (here $1$ m) at the end of an episode. The start and goal locations are randomly sampled at opposite ends of the environment for each episode. Crashes are defined as instances where a robot is in contact with another object in the environment, this environment is then immediately reset. A timeout is said to happen when the robot has remained collision-free till the end of an episode but has not reached the goal location. The parameters of the reward function are varied linearly as the curriculum level increases such that $\lambda^{n}_{i} = \kappa_i n$, and $\eta^{n}_{i} = \xi_i n$, where $\lambda^{n}_{i}$ and $\eta^{n}_{i}$ are the values of $\lambda_i$ and $\eta_i$ at curriculum level $n$, while $\kappa_i$ and $\xi_i$ are positive constants.

To make the network robust against real-world uncertainty, random forces and torques are applied to the simulated robot at discrete time instances sampled from a Bernoulli distribution. Additionally, the position and the orientation of the camera are perturbed by small values (between $\pm5$ cm and $\pm3$ deg). The observations are also perturbed by small values to approximately simulate the uncertainty from real-world sensor measurements. To enable fast learning, we limit the image capture rate to approximately $10$ Hz, while the physics simulation occurs at $100$ Hz in simulated time. To simulate real-world sensor latency, we vary the number of timesteps simulated by the physics engine between two sensor measurements. We add Gaussian noise to the simulated depth images by sampling from a distribution with the standard deviation linearly dependent on the depth value of the pixel. Importantly, in Aerial Gym the dynamics of the simulated agent is matched with those of the real multirotor vehicle. Moreover, the parameters of the velocity controller are randomized to vary the step-response time constant of the robot by $\pm 10\%$.


\section{Evaluation Studies}\label{sec:evaluation}
The proposed approach is extensively evaluated both in simulation and experimental studies.

\subsection{Simulation Studies}

Two sets of simulation studies are conducted to evaluate the method prior to experimental deployment. First, an extensive set of results are derived using the Isaac Gym-powered Aerial Gym simulator. Second, to test the method in a different simulation environment and further deploy it in virtual environments that are particularly different from those experienced in training some indicative results are recorded using Flightmare~\cite{song2020flightmare}. 

\subsubsection{Aerial Gym-based Evaluation}

\begin{figure}[ht!]
\centering
    \includegraphics[width=0.99\columnwidth]{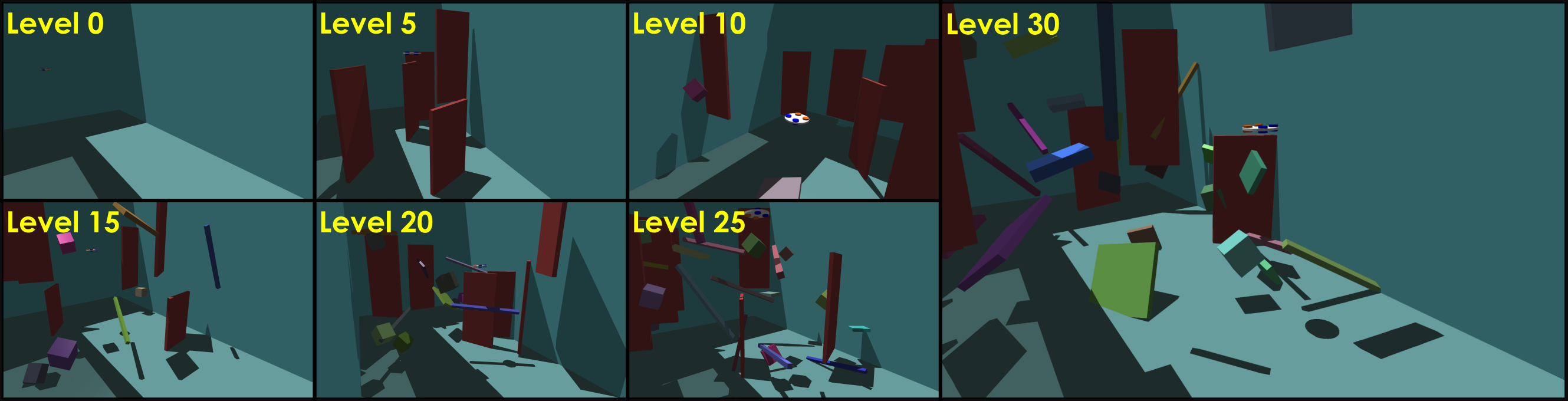}
\vspace{-3ex} 
\caption{Indicative simulation studies using the Aerial Gym simulator to evaluate the trained policy against increasingly complex environments.}
\label{fig:aerialgymenvs}
\vspace{-1ex} 
\end{figure}

The method was first evaluated in Aerial Gym. Specifically, considering $30$ different curriculum levels, representing different levels of complexity in terms of obstacle clutter, the policy was tested for $500$ runs per curriculum level. Each of the tests involved a box-shaped environment with dimensions $L \times W \times H$ within the set $[8,12] \times [5,8], \times [4,6]$. The curriculum level $n$ implies the number of obstacles in the environment (i.e., curriculum level 0 implies 0 obstacles, and curriculum level $30$ implies $30$ obstacles). From curriculum level 0 to level 5 the obstacles are large panels, while after the $5$-th level increasing the level complexity is done through small obstacles employing primitive shapes such as boxes. Figure~\ref{fig:aerialgymenvs} presents indicative environments from different curriculum levels, while Table~\ref{tab:aerialgymresults} summarizes the performance the policy is achieving across complexity levels. The maximum speed and yaw-rate were set to $s_{\textrm{max}} = 1.5\textrm{m/s}$ and $\omega_{\textrm{max}}=60~\textrm{deg/s}$. For environments up to a certain amount of obstacles, the performance is high but it naturally drops in settings with higher clutter.

\begin{table}[ht!]
\centering
\caption{\label{tab:aerialgymresults}Evaluation of the trained policy against environments of different complexity.}
\begin{tabular}{ |c|c|c|c| }
\hline
 Level & Success \% & Timeout \% & Crash \% \\
 \hline
 0 & 99.4 & 0.5 & 0.0\\ 
 5 & 92.8 & 2.1 & 5.0\\ 
 10 & 90.6 & 2.7 & 6.6\\
 15 & 86.3 & 3.7 & 9.9 \\
 20 & 78.5 & 4.4 & 17.0 \\
 25 & 72.7 & 8.1 & 19.1 \\
 30 & 70.8 & 7.2 & 21.8 \\
 \hline
\end{tabular}
\vspace{-4ex}
\end{table}


\subsubsection{Case Studies in Flightmare}

\begin{figure}[ht!]
\centering
    \includegraphics[width=0.99\columnwidth]{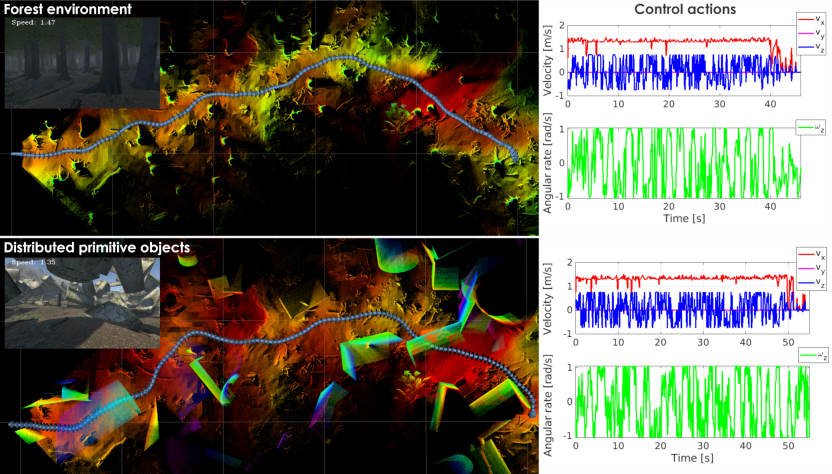}
\vspace{-5ex} 
\caption{Simulation studies using the Flightmare simulator with the goal of evaluating the method given environment diversity --compared to training data-- especially in the case of the forest. On the right the commanded velocities (blue $v_z$, magenta $v_y$ which is zero and $v_x$ and yaw rate $\omega_z$ (green) are shown.} 
\label{fig:vae_rl_flightmare}
\vspace{-3ex} 
\end{figure}

Two case studies were ran in Flightmare and presented in Figure~\ref{fig:vae_rl_flightmare}. Specifically, the method is deployed using the Hummingbird aerial vehicle model ferrying a depth sensor as described in~\cite{Loquercio2021Science} and deployed inside a) a forest, with density governed by a Poisson disc radius of $4\textrm{m}$ and b) an environment with cylinder and cube objects distributed according to a poisson radius of $6\textrm{m}$. It is highlighted that neither type of objects, and especially trees were experienced during the training of the policy or of the \ac{dce}. For both tests the robot is flying is a maximum speed of $s_{\textrm{max}} = 1.5\textrm{m/s}$ and $\omega_{\textrm{max}}=60~\textrm{deg/s}$.

\subsection{Experimental Studies}

The proposed approach was evaluated in experiments involving navigation through cluttered environments with the goal of assessing the overall performance and especially how the sim2real gap is handled. Of special interest was to assess the emerging behaviors including the robot opting to fly above all obstacles when allowed and possible in the environment, alongside its ability to maneuver through clutter when necessary. 

\subsubsection{System Overview}

\begin{figure}[ht!]
\centering
    \includegraphics[width=0.96\columnwidth]{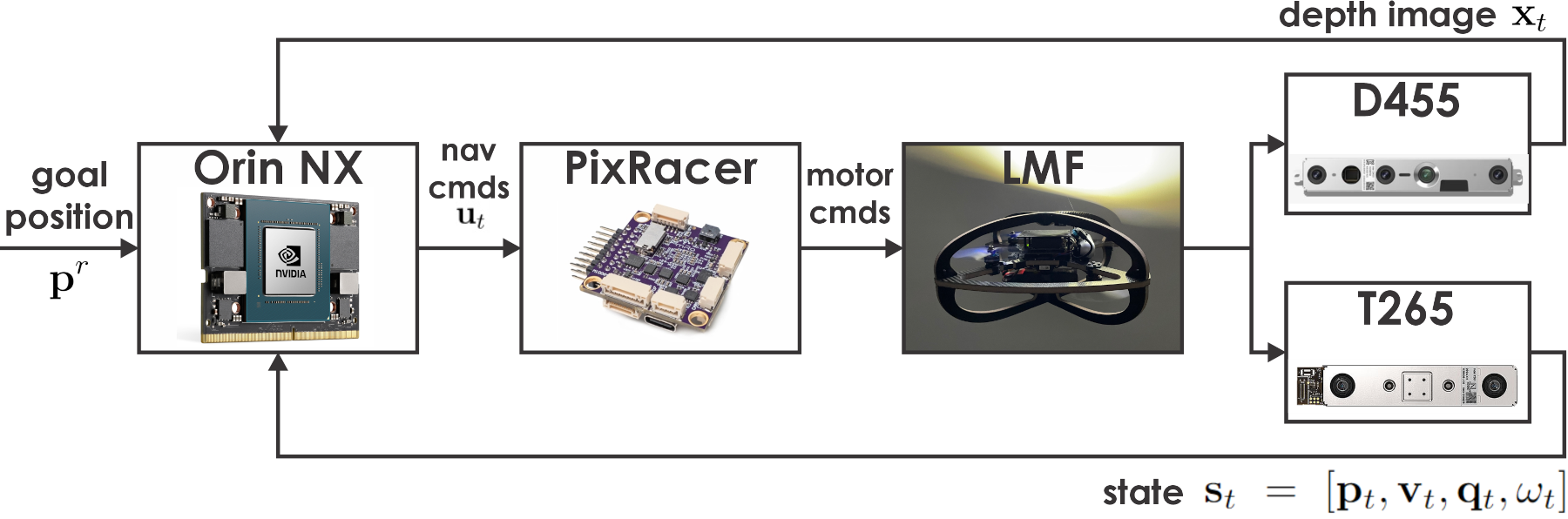}
\caption{Block diagram of the \ac{lmf} robot. The proposed method is implemented onboard the Orin NX processor, while low-level commands are tracked by the embedded autopilot. A Realsense D455 sensors is providing the depth images $\mathbf{o}_t$, while a T265 camera offers odometry estimates $\mathbf{s}_t$. }
\label{fig:lmf_hardware}
\vspace{-2ex} 
\end{figure}

To evaluate the method, we utilize a quadrotor called \ac{lmf}, evolved out of the works in~\cite{Paolo2020RMF} and~\cite{ORACLE,sevae_oracle}. The robot has a diameter of $0.43\textrm{m}$ and weighs $1.2\textrm{kg}$. It features a Realsense D455 for depth and RGB data at $640\times 480$ resolution and $15$ FPS, a PixRacer Ardupilot-based autopilot for velocity and yaw-rate control, and a Realsense T265 fused with the autopilot's \ac{imu} for acquiring the robot's odometry state estimate. The system integrates an NVIDIA Orin NX in which the proposed method is executed exploiting its GPU. The platform is depicted in Figure~\ref{fig:lmf_hardware}.

\begin{figure*}
\centering
    \includegraphics[width=0.99\textwidth]{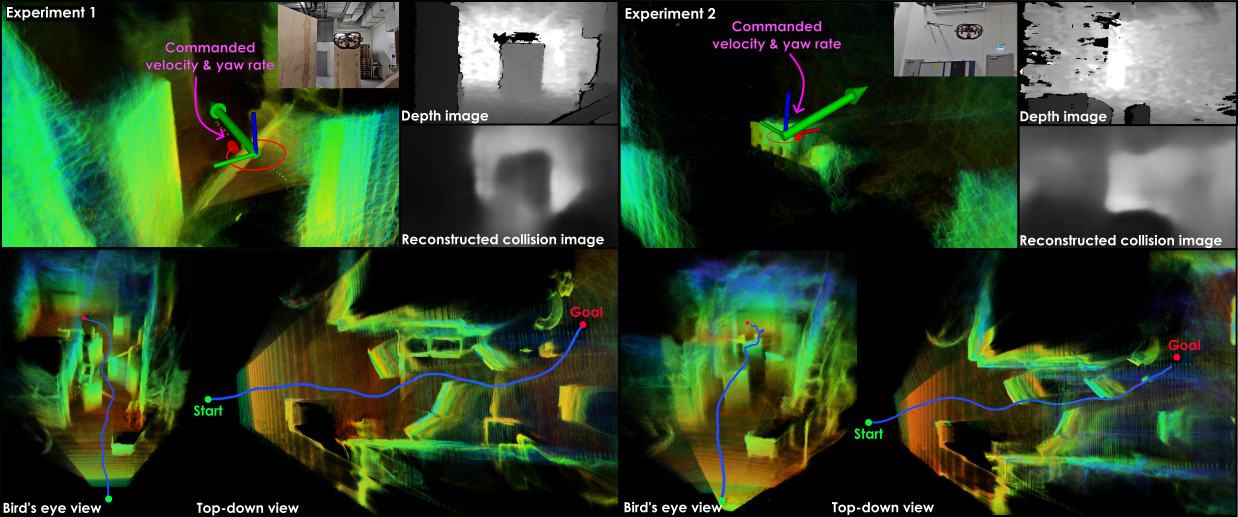}
\caption{Experiments in a cluttered corridor where the vertical velocity of the robot is constrained (in Experiment 1) and matched with training value (in Experiment 2) show that the robot is able to negotiate cluttered environments both passing safely through them when the vertical velocity is constrained, and going over them when possible along the shorter path to the goal location, exploiting the free space above the obstacles. The reconstructions from the \ac{dce} show that the encoder network is robust to noise and imperfect depth data from real-world sensors.}
\label{fig:clutteredtest}
\vspace{-2ex}
\end{figure*}

\begin{figure}[ht!]
\centering
    \includegraphics[width=0.99\columnwidth]{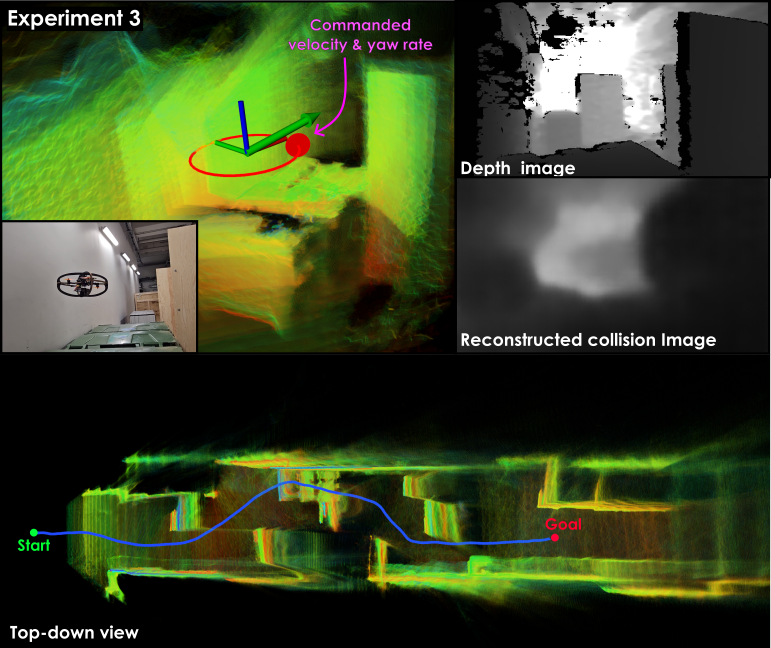}
\caption{Experimental evaluation in a longer cluttered corridor shows that the policy guides the robot safely to the goal location. The robot takes a more intuitive (and shorter) path to the destination choosing to travel through a more narrow opening. The method is robust to the imperfect depth reconstructions owing to the light sources in the environment as seen in the reconstructed collision image.}
\label{fig:cluttered_test_long}
\vspace{-4ex} 
\end{figure}

\subsubsection{Flying through or above cluttered settings}

First, the system is deployed in a cluttered environment, depicted in Figure~\ref{fig:clutteredtest}, and commanded to fly to a waypoint that is $15\textrm{m}$ forward compared to the starting location, $3\textrm{m}$ to the left and at a height of $1\textrm{m}$. Two variations of the experiment are ran, namely a) one in which the vertical velocities of the robot are constrained (with $i_{\textrm{max}} = 7.5~\textrm{deg}$) to prevent its ability to fly above the obstacles, and b) one in which the vertical velocities constraint is removed (with $i_{\textrm{max}} = 30~\textrm{deg}$). As shown in Figure~\ref{fig:clutteredtest} the robot is able to negotiate the environment in both cases. When allowed to exploit its vertical navigation capabilities it acts in the intuitive manner of flying above all obstacles, while when it can only fly by maneuvering around all obstacles it accelerates and decelerates as necessary to achieve so. Importantly, this environment is denser than those encountered in training, so the maximum allowed speed and yaw-rate are set to $s_{\textrm{max}}=1.2\textrm{m/s}$ and $\omega_{\textrm{max}} = 40~\textrm{deg/s}$ for both experiments.

\subsubsection{Maneuvering in a cluttered corridor}
 Finally, we deploy the system in long corridor cluttered with obstacles depicted in Figure~\ref{fig:cluttered_test_long}. The robot is commanded to reach a location that is $20\textrm{m}$ forward and at a height of $1\textrm{m}$. Similar to the previous experiment, we limit the velocity of the robot to $1.2 \textrm{m/s}$, the maximum yaw-rate to $\omega_{\textrm{max}} = 40~\textrm{deg/s}$ and $i_{\textrm{max}} = 7.5~\textrm{deg}$. The robot navigates the clutter and reaches the goal location.

 The depth images during these experiments contained sensor noise and imperfect depth estimation resulting in invalid pixels. However, the reconstructions show that the \ac{dce} is robust to such noise and the encoded latent representation does not affect the performance of the RL policy that has been trained only in simulation using simulated depth images. Furthermore, the obstacles are novel - compared to the training set - which is another testament for its robust sim2real transfer primarily attributed to the modularized architecture and the low-dimensional encoding of collision information by the \ac{dce}. Finally, the proposed method has an inference time of $15~\textrm{ms}$ on the NVIDIA Orin NX board without any optimizations for accelerating inference.

\section{Conclusions}\label{sec:concl}
This paper presented a \ac{drl}-based navigation policy trained completely in simulation, that exploits highly compressed latent representation of depth images obtained from a deep collision encoder network that is trained on both simulated and real data. Extensive evaluation studies conducted in simulation against different obstacle densities and also in different environments show that the method is robust to unseen environments and variations in robot dynamics. Furthermore, real-world experiments show that the method is robust to sensor noise and can navigate the robot safely in cluttered environments.




\bibliographystyle{IEEEtran}
\bibliography{BIB/VAE_RL_ICRA_2024.bib}

\end{document}